# *Biomedical knowledge graph-optimized prompt generation for large language models*

Karthik Soman[1], Peter W Rose[2], John H Morris[3], Rabia E Akbas[1], Brett Smith[4], Braian Peetoom[1], Catalina Villouta-Reyes[1], Gabriel Cerono[1], Yongmei Shi[5], Angela Rizk-Jackson[5], Sharat Israni[5], Charlotte A Nelson[6], Sui Huang[4], Sergio E Baranzini[1*]

[1] Department of Neurology. Weill Institute for Neurosciences. University of California San Francisco. San Francisco, CA, USA.

[2] San Diego Supercomputer Center, University of California, San Diego, La Jolla, CA, USA.

[3] Department of Pharmaceutical Chemistry, School of Pharmacy, University of California, San Francisco, San Francisco, CA, USA.

[4] Institute for Systems Biology, Seattle, WA, USA.

[5] Bakar Computational Health Sciences Institute, University of California, San Francisco, San Francisco, CA, USA.

[6] Mate Bioservices, Inc. Swallowtail Ct. Brisbane, CA, USA.

Short title: Empowering large language models using biomedical knowledge graph

* Corresponding author

E-mail: sergio.baranzini@ucsf.edu (SEB)

# Abstract

Large Language Models (LLMs) are being adopted at an unprecedented rate, yet still face challenges in knowledge-intensive domains like biomedicine. Solutions such as pre-training and domain-specific fine-tuning add substantial computational overhead, requiring further domain-expertise. Here, we introduce a token-optimized and robust Knowledge Graph-based Retrieval Augmented Generation (KG-RAG) framework by leveraging a massive biomedical KG (SPOKE) with LLMs such as Llama-2-13b, GPT-3.5-Turbo and GPT-4, to generate meaningful biomedical text rooted in established knowledge. Compared to the existing RAG technique for Knowledge Graphs, the proposed method utilizes minimal graph schema for context extraction and uses embedding methods for context pruning. This optimization in context extraction results in more than 50% reduction in token consumption without compromising the accuracy, making a cost-effective and robust RAG implementation on proprietary LLMs. KG-RAG consistently enhanced the performance of LLMs across diverse biomedical prompts by generating responses rooted in established knowledge, accompanied by accurate provenance and statistical evidence (if available) to substantiate the claims. Further benchmarking on human curated datasets, such as biomedical true/false and multiple-choice questions (MCQ), showed a remarkable 71% boost in the performance of the Llama-2 model on the challenging MCQ dataset, demonstrating the framework's capacity to empower open-source models with fewer parameters for domain-specific questions. Furthermore, KG-RAG enhanced the performance of proprietary GPT models, such as GPT-3.5 and GPT-4. In summary, the proposed framework combines explicit and implicit knowledge of KG and LLM in a token optimized fashion, thus enhancing the adaptability of general-purpose LLMs to tackle domain-specific questions in a cost-effective fashion.

# Introduction

Large language models (LLM) have shown impressive performance in solving complex tasks across various domains that involve language modeling and processing.(Zhao *et al.* 2023) LLMs are pre-trained on a large corpora of text data in a self-supervised learning framework which can be either Masked Language Modeling (e.g. BERT like models)(Devlin *et al.* 2018; Liu *et al.* 2019) or Auto-Regressive framework (GPT like models)(Brown *et al.* 2020; Luo *et al.* 2022). This pre-training encodes knowledge about the language in the model parameters. Similar to the transfer learning approach commonly used in deep neural networks, this implicit knowledge can be refined through supervised training to excel in a range of domain-specific tasks.(Wei *et al.* 2021; Luo *et al.* 2022) Nevertheless, the "implicit representation" of knowledge in LLM has also been shown to generate non-factual information despite linguistically coherent answers (i.e. "hallucination") as a response to the input prompt.(Maynez *et al.* 2020; Raunak, Menezes and Junczys-Dowmunt 2021; Ji *et al.* 2022) This issue poses a significant challenge for the adoption of LLM models in domains with stringent requirements for accuracy, such as biomedicine. Various strategies have been introduced to address hallucinations in LLMs. One such solution involves the utilization of domain-specific data for pre-training the LLM, rather than relying on generic text corpora. This approach has led to the creation of models such as PubMedBERT(Yu Gu Microsoft Research, Redmond, WA *et al.* 2021), BioBERT(Lee *et al.* 2019), BlueBERT(Lee *et al.* 2019; Peng, Yan and Lu 2019), SciBERT(Beltagy, Lo and Cohan 2019), ClinicalBERT(Huang, Altosaar and Ranganath 2019), BioGPT(Luo *et al.* 2022), Med-PaLM(Singhal *et al.* 2023), and BioMedGPT.(Luo *et al.* 2023) However, pre-training an LLM from scratch imposes a significant computational cost and time overhead to attain the desired human-like performance. An alternative approach, known as prompt tuning, was recently proposed as a means to enhance LLM performance, for instance through the use of zero-

shot,(Kojima *et al.* 2022) few-shot,(Brown *et al.* 2020) and Chain-of-Thought(Wei *et al.* 2022b) prompting strategies.

Although prompt tuning methods have proven to be effective, their performance is restricted to knowledge-intensive tasks that require providing provenance and up-to-date knowledge about the world to address the user prompt. To address such knowledge-intensive tasks, an alternative approach which integrates knowledge graphs (KG) with language models was recently introduced.(Lin *et al.* 2019; Lv *et al.* 2019; Wang *et al.* 2019; Yang *et al.* 2019; Feng *et al.* 2020; Yasunaga *et al.* 2021, 2022) This approach was primarily implemented in question-answering tasks, where the structured information contained within the KG was used to provide context for predicting the answer to the question. While such multimodal integrative approach showed promise, its downstream supervised training was tailored to a specific task, limiting its versatility and broad applicability, thereby constraining its ability to fully harness the "emergent capabilities" of LLMs.(Wei *et al.* 2022a) Retrieval-augmented generation (RAG) involves enhancing a parametric pre-trained LLM with the ability to access a non-parametric memory containing updated knowledge about the world (for e.g., Wikipedia or SPOKE)(Lewis *et al.* 2020)

In this paper, we propose a robust and token-optimized framework called KG-RAG that integrates a KG with a pre-trained LLM within a RAG framework, thus capturing the best of both worlds. To achieve this, we make use of the biomedical KG called (SPOKE (Morris *et al.* 2023)) that integrates more than 40 publicly available biomedical knowledge sources of separate domains where each source is centered around a biomedical concept, such as genes, proteins, drugs, compounds, diseases, and one or more of their known relationships. Because these concepts are recurrent entities forming defined sets (e.g., all named human genes, all FDA approved drugs, etc.) the integration of these concepts into a single graph exposes novel multi-hop factual relationships that connect the knowledge sources and provides the biological and ontological context for each concept. Unlike other RAG approaches, the proposed framework performs an optimized retrieval, specifically obtaining only the essential biomedical context from SPOKE, referred to as 'prompt-aware context', which is adequate enough to address the user prompt with accurate provenance and statistical evidence. This enriched prompt is further used as input for the LLM in the RAG framework for meaningful biomedical text generation. We evaluated this approach using various pre-trained LLMs including Llama-2-13b, GPT-3.5-Turbo and GPT-4.

## Results

Here we developed a framework, called KG-RAG, that integrates with the SPOKE knowledge graph to retrieve accurate and trustworthy biomedical context for LLMs in an optimized and cost-effective fashion from a knowledge graph. This framework involves multiple steps, namely: i) entity recognition from user prompt; ii) extraction of biomedical concepts from SPOKE; iii) concept embedding; iv) prompt-aware context generation; v) conversion into natural language; vi) prompt assembly; and vii) answer retrieval. The performance of this approach was extensively tested using different scenarios.

### Prompting KG-RAG framework

Fig. 1 shows two biomedical prompts (Fig 1A and B yellow box) given as input to the GPT-4 model using two approaches: (i) prompt based, i.e., without KG-RAG (blue box) and (ii) with KG-RAG (green box). We observed that only KG-RAG was able to provide an accurate answer for both prompts, accompanied by supporting evidence and provenance information. (For more prompting examples refer Table A in S1 Text. We have also provided these prompts in a Jupyter notebook which can be accessed at https://github.com/BaranziniLab/KG_RAG/blob/main/notebooks/kg_rag_based_gpt_prompts.ipynb)

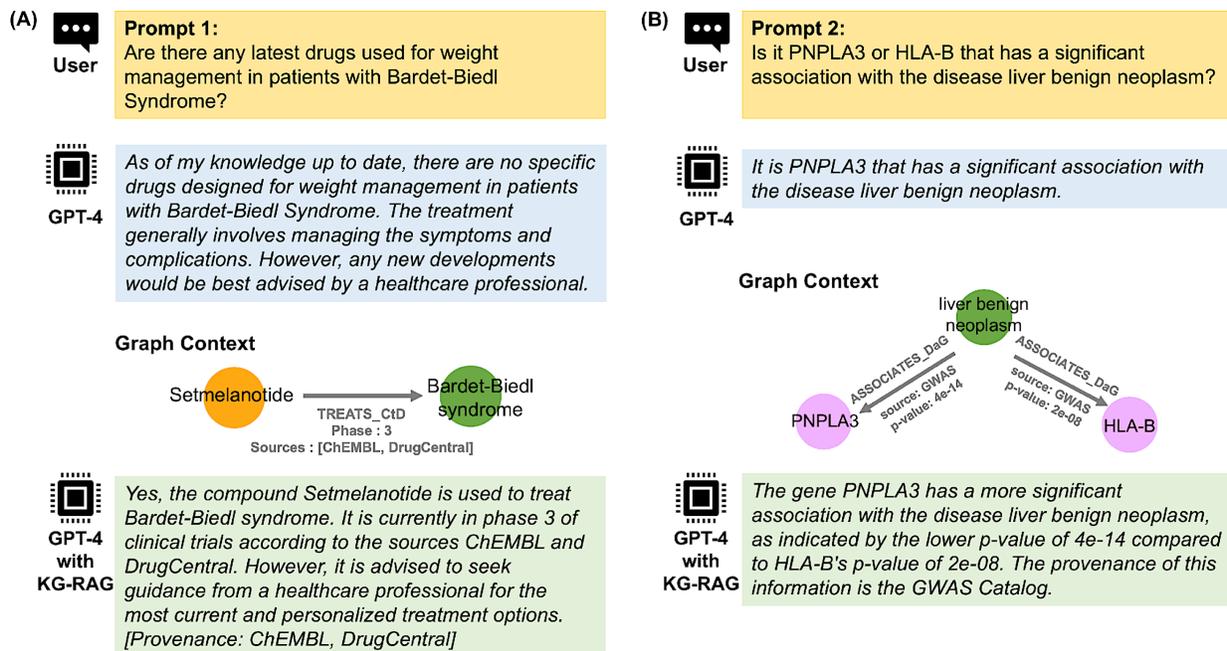

**Fig 1. Prompting examples** Example prompts (yellow boxes) and prompt based (blue boxes)/KG-RAG based (green boxes) responses using GPT-4. **(A)** shows a biomedical prompt to search the drugs used for weight management in patients with 'Bardet-Biedl Syndrome' and **(B)** shows another prompt to compare the statistical associations of PNPLA3 and HLA-B genes with the disease liver benign neoplasm. Both (A) and (B) show the associated 'prompt aware graph' context.

# Hyperparameter analysis

KG-RAG has two hyperparameters, ('context volume' and 'context embedding model'), which enable it to conduct optimized context retrieval from a KG. Context volume defines the upper limit on the number of graph connections permitted to flow from the KG to the LLM (See Materials and methods, S1 Text). Context embedding model extracts graph context that shows semantic similarity to the user prompt, facilitating the refinement of extracted context to those that are contextually relevant (See Materials and methods, S1 Text). To optimize these hyperparameters, we used two context embedding models (MiniLM and PubMedBert) with increasing sizes of context volume (Fig 2A). For prompts with single disease entity, the PubMedBert based model exhibited a mean performance (Jaccard similarity) approximately 10% higher than that of the MiniLM model across all context volume settings (mean performance of PubMedBert = 0.67, mean performance of MiniLM = 0.61). For prompts with two disease entities, PubMedBert achieved a performance that was 8.1% higher than the MiniLM across all context volume settings (mean performance of PubMedBert model = 0.4, performance of MiniLM = 0.37).

Fig 2A shows that the performance curve reaches a plateau for prompts with single disease entity and follows a similar trend for prompts with two disease entities (for both models). Based on these findings, we selected PubMedBert based model as the context embedding model and set the context volume to a value between 100-200 (For most downstream tasks, we opted for a context volume of 150, and for True/False questions, we chose a context volume of 100).

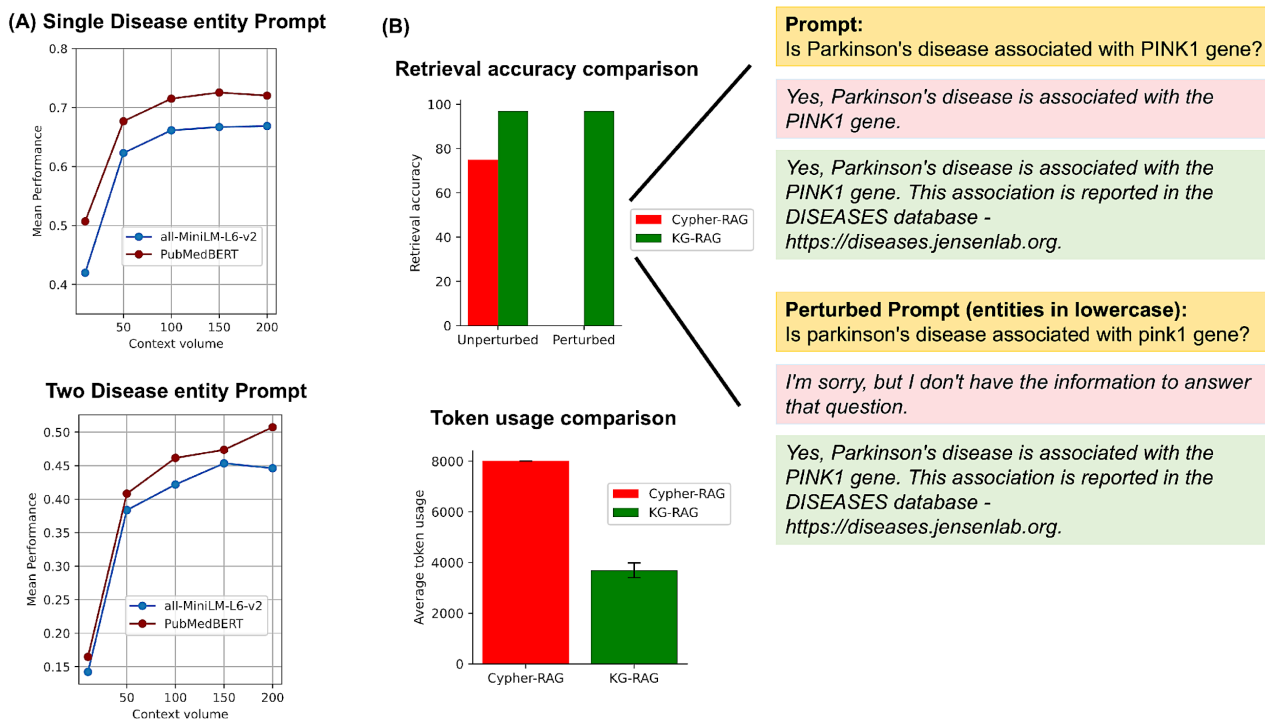

**Fig 2. Hyperparameter analysis and RAG comparison (A)** Hyperparameter analysis performance curves using prompts with single (top) and two (below) disease entities mentioned in it. The x-axis denotes the 'Context volume' (number of associations from KG) and the y-axis denotes the mean performance (Jaccard similarity) across the prompts. The red curve denotes 'PubMedBert' and the blue curve denotes 'MiniLM' transformer models. **(B)** shows the comparative analysis between KG-RAG (green color) and Cypher-RAG (red color) in terms of retrieval accuracy (top) and token usage (bottom). Insight shows an example where Cypher-RAG fails to retrieve context from the KG when the input prompt is slightly perturbed, but KG-RAG remains robust in context retrieval. It is evident that KG-RAG has lesser token usage (average of 53.9% reduction in token usage) when compared to Cypher-RAG (bottom). Error bar in the token utilization bar plot (bottom) represents standard error of the mean (sem).

## RAG comparative analysis

Fig 2B shows the comparative analysis between the proposed KG-RAG and the existing Cypher-RAG approach for context retrieval from a KG (See Materials and Methods). We compared these two frameworks based on their retrieval accuracy, retrieval robustness and token usage. For a test dataset with 100 biomedical questions (S1 Text), Cypher-RAG and KG-RAG showed 75% and 97% retrieval accuracy respectively (Fig 2B top). To test the robustness in context retrieval, we introduced a slight perturbation to the test dataset by converting the entity names to lowercase (Fig 2B insight). We observed a significant decrease in the retrieval accuracy of Cypher-RAG to 0% (indicating failure to retrieve any context from the graph). This mainly occurs because Cypher-RAG utilizes precise matching of the entity keywords provided in the user prompt to formulate the Cypher query for extracting graph context. On the otherhand, KG-RAG maintained its retrieval accuracy at 97% (i.e. robust to the input perturbation, Fig 2B top). This is because KG-RAG employs a semantic embedding approach to extract graph context, which enhances its ability to effectively handle various representations of entities within user prompts. Next, we analyzed the total token usage of each framework for generating the response for the same test dataset (Fig 2B bottom). We found that Cypher-RAG had an average token usage of 8006 tokens whereas KG-RAG had an average token usage of 3693 tokens (Fig 2B bottom). This represents a 53.9% reduction in the token usage by KG-RAG compared to Cypher-RAG which points to a significant cost-effective retrieval ability of KG-RAG.

## Performance on True/False and MCQ datasets

Fig 3 shows bootstrap distributions of performance (accuracy) of the three LLMs using prompt-based and KG-RAG framework on True/False (Fig 3A) and MCQ (Fig 3B) datasets. Table 1 summarizes the performance of the three LLMs across these datasets. We observed a consistent performance enhancement for the LLM models under KG-RAG framework on both True/False and MCQ datasets (Table 1). KG-RAG significantly elevated the performance of Llama-2 by approximately 71% from its initial level (0.31±0.03 to 0.53±0.03) on the more challenging MCQ dataset (Table 1). Intriguingly, we also observed a small but significant drop in the performance of GPT-4 model (0.74±0.03) compared to GPT-3.5-Turbo model (0.79±0.02) on MCQ dataset using KG-RAG framework (T-test, p-value < 0.0001, t-statistic = -47.7, N = 1000) but not in the prompt-based approach.

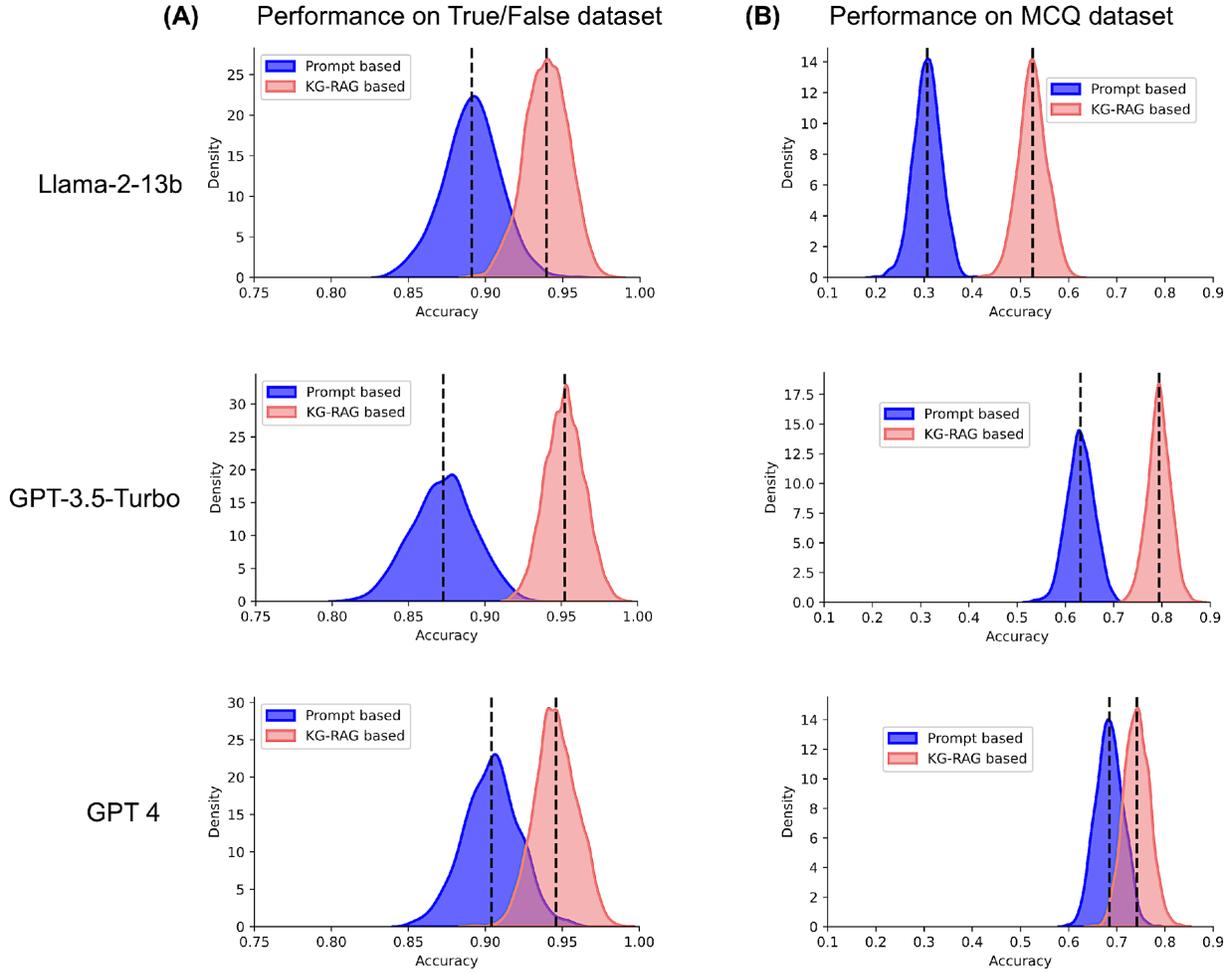

**Fig 3. LLM performance on True/False and MCQ datasets** Performance (Accuracy) distributions of LLMs on **(A)** True/False and **(B)** MCQ datasets. Blue distribution denotes the performance using prompt-based approach and red distribution denotes the performance using KG-RAG based approach. Black vertical dashed line indicates the mean value of the distribution. The higher the value, the better the performance.

**Table 1. LLM performance (Accuracy: mean ± std) on True/False and MCQ dataset**

| Model | True/False dataset | | MCQ dataset | |
|---|---|---|---|---|
| | Prompt-based | KG-RAG | Prompt-based | KG-RAG |
| Llama-2-13b | 0.89±0.02 | 0.94±0.01 | 0.31±0.03 | 0.53±0.03 |
| GPT-3.5-Turbo | 0.87±0.02 | 0.95±0.01 | 0.63±0.03 | 0.79±0.02 |
| GPT-4 | 0.9±0.02 | 0.95±0.01 | 0.68±0.03 | 0.74±0.03 |

# Discussion

In this work, we introduce a simple but highly effective framework that combines a biomedical knowledge graph with LLM chat models in a token optimized fashion. This integration resulted in a domain-specific generative system whose responses were firmly grounded in well-established biomedical knowledge. We compared the proposed framework with another RAG approach that utilizes Cypher query and showed that KG-RAG was more robust to prompt perturbation and more efficient in token utilization. In addition, KG-RAG consistently demonstrated superior performance compared to the prompt-based baseline LLM model on all human-curated benchmark datasets. We hypothesize that this performance improvement arises from the fusion of the explicit knowledge from the KG and the implicit knowledge from the LLM. This shows the value of providing domain-specific ground truth at a fine-grained resolution as context at the prompt level.

A heterogeneous knowledge graph with diverse concepts (the biomedical concepts in this case) interconnected at a massive scale has the potential to generate new knowledge as an "emergent property".(Baranzini *et al.* 2022; Morris *et al.* 2023) In fact, as LLMs scale up in various dimensions like model parameters, training data, and training compute, they have been thought to exhibit reasoning or "emerging capabilities"(Wei *et al.* 2022a) although this observation could also be explained by "in-context learning" or other aspects of the examples.(Brown *et al.* 2020; Min *et al.* 2022; Lu *et al.* 2023) In any case, KG-RAG capitalized this capability and generated biomedical text with rich annotations such as provenance and statistical evidence (if available) thereby resulting in more reliable and knowledge-grounded responses. Additionally, the optimized and fine-grained context retrieval capability of KG-RAG ensured a budget friendly RAG system to apply on proprietary LLMs.

Previous studies have utilized KG in conjunction with LLM for knowledge intensive tasks such as question-answering,(Yasunaga *et al.* 2021) multi-hop relational reasoning,(Feng *et al.* 2020) commonsense reasoning,(Lin *et al.* 2019; Lv *et al.* 2019) and model pre-training.(Moiseev *et al.* 2022; Yasunaga *et al.* 2022) Furthermore, enhancing prompts by incorporating structured knowledge has been described and studied.(Xiang Chen Zhejiang University, China *et al.*; Lewis *et al.* 2020; Pan *et al.* 2023) Naturally, these approaches have bolstered the positive reinforcement between KG and LLM. Nevertheless, it's worth noting that these approaches are only be task-specific and in some cases, the knowledge infusion could grow exponentially by the inclusion of higher order relations.(Bauer, Wang and Bansal 2018; Lin *et al.* 2019) Such approaches could compromise the limited token space of the LLM. Alternative methods employed knowledge infusion through the direct use of query languages like SPARQL.(Brate *et al.* 2022) However, this could render the system constraint to the schema of the underlying KG, potentially affecting the flexibility and adaptability of prompts. Moreover, as the KG expands and its schema grows, it could potentially occupy a significant portion of the LLM input token space. This explains why we noticed a greater token usage with the Cypher-RAG method (average usage of 8006 tokens), as it incorporates the entire graph schema into the input prompt for converting natural language into structured Cypher queries. In contrast, KG-RAG requires minimal graph schema, thus eliminating the need to include it in the prompt and resulting in substantial token savings, with a reduction of over 50% in token utilization compared to Cypher-RAG. This finding suggests that when dealing with a graph as large as SPOKE, which contains over 40 million nodes, Cypher-RAG requires LLMs that enable a larger context window. In contrast, KG-RAG is capable of managing this with LLMs that require a relatively smaller window size.

To conduct robust benchmarking, we curated datasets that underwent review by domain experts. Given the swift progress in LLM research, we believe that such rigorously vetted datasets could serve as valuable resources not only for evaluating KG-RAG but also for assessing other ongoing LLM endeavors in biomedicine. In our benchmarking analysis, we found an enhancement in LLM performance as a function of the model size in terms of the number of parameters. Intriguingly, with the KG-RAG framework the performance of GPT-4 on the MCQ dataset, despite its model size, dropped significantly compared to that of the GPT-3.5-Turbo on the MCQ dataset. In fact, the performance of GPT-3.5-Turbo under KG-RAG framework was on par with that of the GPT-4 model on True/False datasets. These results suggest that at present, GPT-3.5 may be a better context listener than GPT-4. In fact, a recent study compared the March 2023 version of GPT-4 with the June 2023 version, shedding light on the drift in the LLM performance over time.(Chen, Zaharia and Zou 2023) The study revealed that, as time progressed, GPT-4 exhibited a reduced tendency to adhere to user instructions.(Chen, Zaharia and Zou 2023) In contrast, there was no consistent alteration observed in the instruction-following behavior of GPT-3.5 over time.(Chen, Zaharia and Zou 2023)

When the proprietary GPT models were compared to the open-source Llama-2-13b model, they showed a narrow margin in performance on the biomedical True/False dataset. However, on the more challenging MCQ dataset, Llama-2 initially demonstrated lower performance compared to GPT models. Interestingly, the KG-RAG framework provided a substantial performance boost to Llama-2, improving its performance by ~71% from the baseline. Despite this boost that narrowed the performance gap, the performance of Llama-2 remained lower than that of the GPT models. This suggests that the KG-RAG framework has the potential to capitalize the intrinsic context comprehension capabilities of open-source pretrained models like Llama-2, making them more competitive with proprietary models like GPT. The findings underscore the importance of context enrichment techniques for improving the performance of language models on complex tasks in specialized domains.

While the proposed framework has successfully addressed numerous challenges, we recognize there are opportunities for improvement. Currently, this approach is limited to handling biomedical questions centered around diseases due to our focus on embedding disease concepts from the SPOKE KG during recognition of the biomedical entity from the prompt. Future work could expand this scope by including all biomedical concepts (nodes) in SPOKE and other KG. Since SPOKE contains more than 40 million biomedical nodes, this expansion will enable the KG-RAG framework to address a broader range of biomedical questions and thereby enhance its versatility. Currently, we implemented the KG-RAG framework on the SPOKE biomedical knowledge graph. However, we believe that the framework is general enough to extend to other biomedical KG and even other domains as well. Finally, the quality of the retrieved context relies on the information stored in the underlying graph. In our case, SPOKE utilizes meticulously curated knowledge bases to construct its nodes and edges; however, we do not assert that it is entirely error-free or ready for clinical use. Thus, while SPOKE's reliability has been demonstrated through its successful application in various biomedical contexts.(Himmelstein and Baranzini 2015; Himmelstein *et al.* 2017; Nelson, Butte and Baranzini 2019; Nelson *et al.* 2021a, 2021b; Baranzini *et al.* 2022; Morris *et al.* 2023; Soman *et al.* 2023a, 2023b; Tang *et al.* 2024), it is important to note that this work focused on the creation of a framework rather than on a rigorous and formal evaluation of the KG itself.

In summary, the KG-RAG framework retrieves semantically meaningful context from a knowledge graph using minimal tokens, then combines this explicit knowledge with the parameterized implicit knowledge of an LLM. This knowledge integration results in the generation of domain-specific, reliable and up-to-date meaningful biomedical responses with rich annotations.

# Materials and Methods

## Knowledge Graph based Retrieval Augmented Generation (KG-RAG) Framework

The Schema of the proposed KG-RAG framework is shown in Fig 4. The following sections explain each component of this framework.

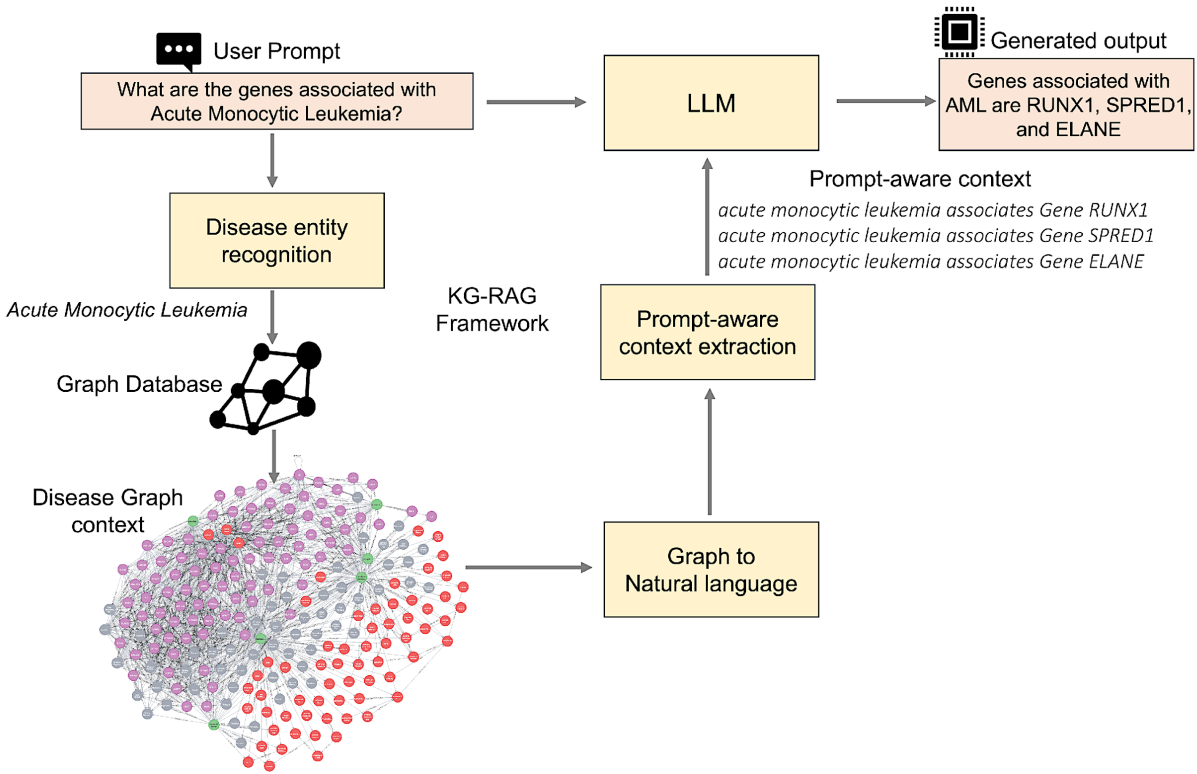

**Fig 4. Schema for the Knowledge Graph based Retrieval-Augmented Generation (KG-RAG) Framework**. The direction of the arrows indicates the flow of the pipeline in this framework

**Disease entity recognition**

This is the first step in KG-RAG. The objective of this step is to extract the disease concept (an entity) from the input text prompt and then find the corresponding matching disease node in the KG (a SPOKE concept). This was implemented as a two-step process: i) entity extraction from prompt and ii) entity matching to SPOKE. Entity extraction identifies and extracts disease entities mentioned in the input text prompt, otherwise called as 'Prompt Disease extraction' (Fig 5). To achieve this, zero-shot prompting(Kojima *et al.* 2022) was used on GPT-3.5-Turbo model (Fig 5). Specifically, a system prompt was designed to extract disease entities from the input text and return them in JSON format (S1 Text).

Next, entity matching was used to obtain the concept name of the disease as it is represented within the KG. For this, the embeddings for all disease concepts (i.e., nodes) in SPOKE were precomputed using the 'all-MiniLM-L6-v2' sentence transformer model (i.e. Disease Latent space in Fig 5).(Reimers and Gurevych 2019) This procedure translates names of the disease concepts into a 384-dimensional dense vector space, making it suitable for semantic search. We chose 'MiniLM' for two main reasons: firstly, when combined with entity extraction, it successfully retrieved the disease nodes from the graph with an accuracy of 99.7% (S1 Text). Second, it produced lightweight embeddings (384 dimensions) compared to other sentence transformers like the PubmedBERT model(Yu Gu Microsoft Research, Redmond, WA *et al.* 2021) (768 dimensions), making it more memory efficient. Next, these newly created disease concept embeddings were stored in the 'Chroma' vector database.(Dhungana 2023) Disease concepts with the highest vector similarity to the extracted entity are selected for subsequent context retrieval (Fig 5). If the zero-shot approach fails to identify a disease entity in the prompt, five disease concepts from the vector database with the most significant semantic similarity to the entire input text prompt are selected instead.

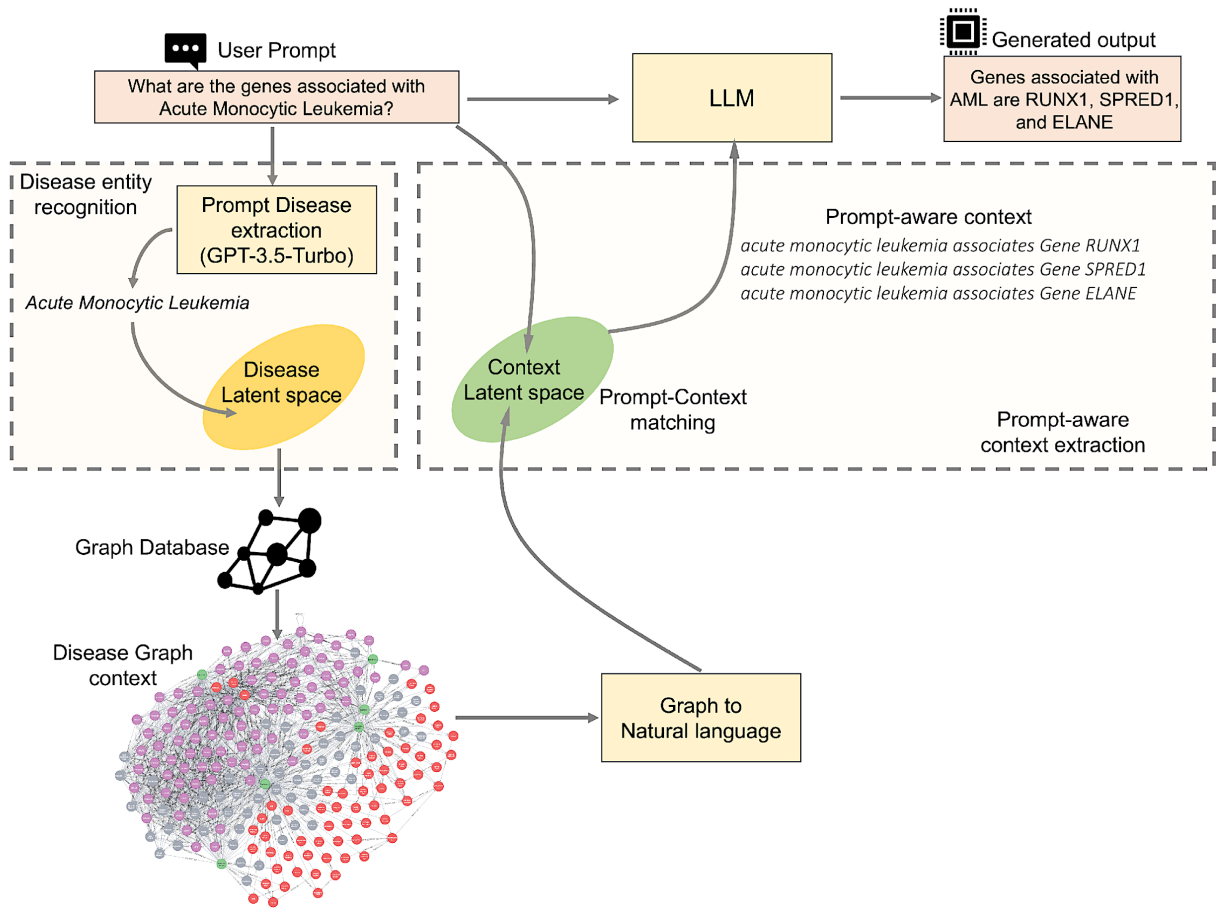

**Fig 5. Detailed schema of KG-RAG.** Dashed boxes show the details of 'Disease entity recognition' and 'Prompt-aware context extraction' from the Knowledge Graph

## Disease context retrieval from SPOKE

The SPOKE KG connects millions of biomedical concepts through semantically meaningful relationships.(Morris *et al.* 2023) The KG consists of 42 million nodes of 28 different types and 160 million edges of 91 types, implemented as a property graph and is assembled by downloading and integrating information from 41 different biomedical databases.(Morris *et al.* 2023) Notably, the vast majority of SPOKE is composed of curated information determined by systematic experimental measurements, not text mining from the literature. In this study, SPOKE was used as the source of biomedical context for the diseases mentioned in the input prompt. In short, the context triples (Subject, Predicate, Object) associated with a disease node from SPOKE KG (Morris *et al.* 2023) were fetched, and subsequently transformed into English language, making it compatible for input into the LLM (Fig 4) (S1 Text). The predicate schema of SPOKE is as follows (S1 Text):

```
upperCase(predicateName)_<upperCase(firstLetter(subjectType)),
lowerCase(firstLetter(predicateName)),
upperCase(firstLetter(objectType))>
```

This schema allowed for the direct conversion of the extracted triples into English language using the following rule (S1 Text):

```
(S, P, O) → Subject lowerCase(predicateName) Object
```

For example:

```
(Disease hypertension, ASSOCIATES_DaG, Gene VHL) → `Disease
hypertension associates Gene VHL`
```

In addition to extracting the connectivity between the disease and its neighbors, we also extracted the provenance information associated with those edges. In SPOKE, provenance of an association is given as its edge attribute. This provenance information is then appended to the respective context. In addition to provenance, edges within SPOKE also have other attributes such as evidence supporting the assertion (e.g., p-value, z-score, enrichment score, etc.). Additionally, we have implemented an option for users to incorporate this evidence information by utilizing a '-e' command line option, which accepts a Boolean value during the execution of the KG-RAG script. If the Boolean is set to True (by default it is set to False), the evidence information will be extracted and appended to the provenance information of the respective context.

**Context Pruning**

Next, the extracted disease context was pruned by selecting the most semantically pertinent ones required to respond to the given prompt as explained below. First, the input prompt and all the extracted contextual associations were embedded to the same vector space (Context Latent space in Fig 5) using a sentence transformer model (model selection was done using hyperparameter tuning). Next, only those contextual associations showing the highest cosine similarity with the input prompt vector were selected (Fig 5). For context selection, prompt-context cosine similarity should satisfy two conditions: (i) greater than 75$^{th}$ percentile of the similarity distribution encompassing all the context related to the chosen disease node and (ii) having a minimum similarity value of 0.5. This makes the retrieved context more fine-grained and contextually relevant.

**Large Language Model (LLM)**

The input prompt, when combined with the prompt-aware context, resulted in an enriched prompt that was used as input to the LLM for text generation. For that purpose, three pre-trained chat models were used: Llama-2-13b,(Touvron *et al.* 2023) GPT-3.5-Turbo and GPT-4.(Brown *et al.* 2020) A Llama model with 13 billion parameters and with a token size of 4096 was downloaded and deployed in the Amazon Elastic Compute Cloud (EC2) GPU P3 instance. GPT models were accessed using the OpenAI API. Since GPT models featured a higher parameter count in comparison to Llama, this gave us the opportunity to compare the performance of KG-RAG as a function of the size of the LLM in terms of its parameter count. In this study, the 'temperature' parameter (Brown *et al.* 2020), governing the level of randomness in the LLM output, was set to 0 for all LLMs.

## Hyperparameter analysis and validation

The performance of KG-RAG was evaluated across two sets of hyperparameters such as 'Context volume' and 'Context embedding model'. Context volume defines the upper limit on the number of graph connections permitted to flow from the KG to the LLM. This hyperparameter introduced a balance between context enhancement and input token space utilization of the LLM (S1 Text). Next hyperparameter called 'Context embedding model' determined which model exhibited greater proficiency in retrieving the accurate biomedical context from the KG to respond to the input prompt. Since 'MiniLM'(Reimers and Gurevych 2019) was used in the disease entity recognition stage, we considered that as a candidate for 'Context embedding model'. Acknowledging that biomedical contexts often utilize vocabulary that could differ from general domain scenarios we next considered 'PubMedBert'(Yu Gu Microsoft Research, Redmond, WA *et al.* 2021; Deka, Jurek-Loughrey and Deepak 2022) as another candidate for this hyperparameter since it was pre-trained on biomedical text. Additionally, two sets of validation data (total 165 questions) were created using prompts mentioning a single disease (75 questions), while the other set involved prompts mentioning two diseases (90 questions) (S1 Text). The second set of 'two disease prompts' were created to see if KG-RAG had the ability to retrieve contexts related to multiple diseases. These prompts were executed using the GPT-4 model in the KG-RAG framework, utilizing specific system prompts to return the results in JSON format (S1 Text). Jaccard similarity was computed by parsing these JSON responses and comparing them with the ground truth. Through these analyses, an empirical selection of hyperparameters for the downstream tasks was made.

## Test dataset

Three types of test datasets were used to quantitatively analyze the performance of the proposed framework (Table B in S1 Text): (i) True/False dataset; (ii) Multiple Choice Question (MCQ) dataset; and (iii) RAG comparison dataset. Datasets i and ii were consolidated from external knowledge bases and were further thoroughly reviewed by domain experts to remove any false positives.

A True/False dataset was created from three external data sources, such as DisGeNET,(Piñero *et al.* 2016) MONDO,(Vasilevsky *et al.* 2022) and SemMedDB.(Kilicoglu *et al.* 2012) DisGeNET consolidates data about genes and genetic variants linked to human diseases from curated repositories, the GWAS catalog, animal models, and the scientific literature(Piñero *et al.* 2016) (S1 Text). MONDO provides information about the ontological classification of Disease entities in the Open Biomedical Ontologies (OBO) format(Vasilevsky *et al.* 2022) (S1 Text). SemMedDB contains semantic predications extracted from PubMed citations(Kilicoglu *et al.* 2012) and we used this resource to formulate True/False questions about drugs and diseases (S1 Text). MCQ comprising five choices with a single correct answer for each question were created using data from the Monarch Initiative(Mungall *et al.* 2017) and ROBOKOP (Reasoning Over Biomedical Objects linked in Knowledge-Oriented Pathways)(Bizon *et al.* 2019) (S1 Text). To assess the graph context retrieval capabilities of various RAG frameworks, we extracted Disease-Gene associations from the SPOKE graph and designed questions based on these associations. These questions were then used on KG-RAG and Cypher-RAG frameworks to compare how well they retrieve the associated context from SPOKE graph.

Thus, 311 True/False, 306 MCQ and 100 RAG comparison biomedical question datasets were created for a systematic quantitative analysis of the proposed framework. To assess the performance of LLMs on True/False and MCQ datasets, 150 questions were randomly sampled with replacement 1000 times (using bootstrapping). The accuracy metric was then calculated for each sampling iteration, resulting in a performance distribution.

## Cypher-RAG

Cypher-RAG is a technique utilized for retrieving context associated with a node in a Neo4j graph database.(Bratanic 2023) This context can then be leveraged to generate information about the node in natural language using a LLM. The method involves explicitly embedding the schema of the graph into the input prompt, directing the LLM to generate a structured Cypher query based on this schema. The resulting Cypher query is used to make a call to the Neo4j database, and the returned information is utilized as context to respond to the user's prompt. This methodology is integrated into the LangChain python library as *GraphCypherQAChain* class.(Bratanic 2023) An advantage of this approach is that it allows for the creation of Cypher queries directly from natural language, eliminating the need for users to have the knowledge of Cypher query syntax. However, our analysis revealed certain limitations of this approach. We found that the explicit embedding of the graph schema restricts the input token space and increases token usage for this method. As the complexity of the graph schema increases, users may need to utilize LLMs with longer context window sizes for optimal performance. Additionally, we demonstrated that this method can be sensitive to how the prompt is formulated. Even slight perturbations to the prompt can lead to incorrect Cypher queries and subsequently impact downstream generative processes.

## Acknowledgements

SEB holds the Heidrich Family and Friends Endowed Chair of Neurology at UCSF. SEB holds the Distinguished Professorship in Neurology I at UCSF. We also acknowledge the Versa team at UCSF for providing the infrastructure to access OpenAI API. The development of SPOKE and its applications are being funded by grants from the National Science Foundation [NSF_2033569], NIH/NCATS [NIH_NOA_1OT2TR003450].
## Data availability

SPOKE KG can be accessed at https://spoke.rbvi.ucsf.edu/neighborhood.html. It can also be accessed using REST-API (https://spoke.rbvi.ucsf.edu/swagger/). KG-RAG code is made available at https://github.com/BaranziniLab/KG_RAG. Biomedical benchmark datasets used in this study are made available to the research community in the same GitHub repository.

# Supporting information

S1 Text : Supplementary information about Context extraction from SPOKE, Hyperparameter analysis, Disease entity recognition retrieval analysis, Validation data creation, Test data creation, System prompts used and two Supplementary tables: Table A displays additional biomedical prompting examples and Table B consolidates sources for the test dataset used in this study.

Supporting information can be accessed at:

https://ucsf.box.com/s/9qlwck02s3jkj9gzpanjzodfpazb7k2q